\documentclass[]{article}
\usepackage{proceed2e}
\usepackage{latexsym}

\title{Programming with Personalized PageRank: \\
A Locally Groundable First-Order Probabilistic Logic} 

\author{} % LEAVE BLANK FOR ORIGINAL SUBMISSION.
\author{ {\bf William Y. Wang} \\  
Language Technology Inst. \\  
Carnegie Mellon University\\ 
Pittsburgh, PA 15213 \\ 
\And 
{\bf Kathryn Mazaitis}  \\ 
Machine Learning Dept.\\
Carnegie Mellon University\\ 
Pittsburgh, PA 15213 \\ 
\And 
{\bf William W. Cohen}   \\ 
Language Technology Inst.\\ 
and Machine Learning Dept.\\
Carnegie Mellon University\\ 
Pittsburgh, PA 15213 \\ 
} 
 
\newcommand{\wc}[1]{} 
\newcommand{\trm}[1]{\emph{#1}}

\newcommand{\uncite}[1]{}
\newtheorem{theorem}{Theorem}
\newtheorem{corollary}{Corollary}
\newcommand{\weightvec}{\textbf{w}}
\newcommand{\edge}[2]{{u\rightarrow{}v}}
\newcommand{\edgeuv}{{\edge{u}{v}}}

\usepackage{epsfig}

\begin{document} 
 
\maketitle 
 
\begin{abstract} 
In many probabilistic first-order representation systems, inference is
performed by ``grounding''---i.e., mapping it to a propositional
representation, and then performing propositional inference.  With a
large database of facts, groundings can be very large, making
inference and learning computationally expensive.  Here we present a
first-order probabilistic language which is well-suited to approximate
``local'' grounding: every query $Q$ can be approximately grounded
with a small graph.  The language is an extension of stochastic logic
programs where inference is performed by a variant of personalized
PageRank.  Experimentally, we show that the approach performs well
without weight learning on an entity resolution task; that supervised
weight-learning improves accuracy; and that grounding time is
independent of DB size.  We also show that order-of-magnitude speedups
are possible by parallelizing learning.
\end{abstract} 
 
\section{INTRODUCTION} 
 
\begin{figure}
\centerline{\includegraphics[scale=0.25]{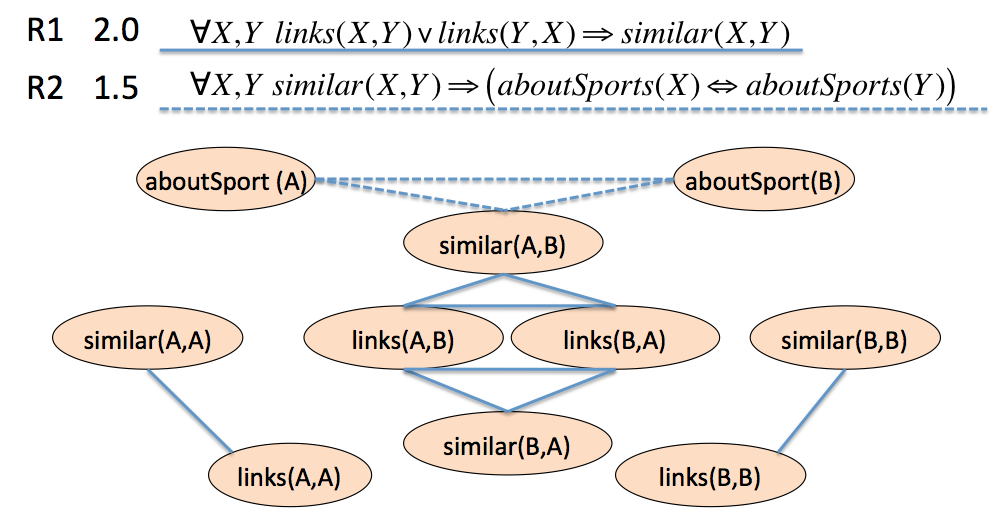}}
\caption{A Markov logic network program and its grounding.  (Dotted
  lines are clique potentials associated with rule R2, solid lines
  with rule R1.)}
\label{fig:mln}
\end{figure}

In many probabilistic first-order representation systems, including
Markov Logic Networks \cite{RichardsonMLJ2006} and Probabilistic
Similarity Logic \cite{brocheler2012probabilistic}, inference is
performed by mapping a first-order program to a propositional
representation, and performing inference in that propositional
representation.  This mapping is often called \trm{grounding}.  For
example, Figure~\ref{fig:mln} shows a simple MLN.\footnote{This MLN
  does a very simple sort of label-propagation through hyperlinks.}
As is often the case, this MLN has two parts: the \trm{rules}
$R_1,R_2$, which are weighted first-order clauses; and the
\trm{database} $DB$, which consists of facts (unit clauses) of the
form \textit{links(a,b)} for constants $a,b$.  The figure also shows
the the grounded version of this MLN, which is an ordinary Markov
network: the DB facts become constraints on node values, and the
clauses become clique potentials.

Grounding a first-order program can be an expensive operation.  For a
realistic hyperlink graph, a Markov network with size even linear in
the number of facts in the database, $|DB|$, is impractically large
for inference.  Superficially, it would seem that groundings must
inheritly be $o(|DB|)$ for some programs: in the example, for
instance, the probability of \textit{aboutSport(x)} must depends to
some extent on the entire hyperlink graph (if it is fully connected).
However, it also seems intuitive that if we are interested in
inferring information about a specific page---say, the probability of
\textit{aboutSport(d1)}--then the parts of the network only distantly
connected to \textit{d1} are likely to have a small influence.  This
suggests that an \emph{approximate} grounding strategy might be
feasible, in which a query such as \textit{aboutSport(d1)} would be
grounded by constructing a small subgraph of the full network,
followed by inference on this small ``locally grounded'' subgraph.
Likewise, consider learning (e.g., from a set of queries $Q$ with
their desired truth values). Learning might proceed by
locally-grounding every query goal, allowing learning to also take
less than $O(|DB|)$ time.

In this paper, we present a first-order probabilistic language which
is well-suited to approximate ``local'' grounding.  We present an
extension to \trm{stochastic logic programs} (SLP)
\cite{DBLP:journals/ml/Cussens01} that is biased towards short
derivations, and show that this is related to \trm{personalized
  PageRank} (PPR) \cite{pagerank,SoumenWWW2007} on a linearized
version of the proof space. Based on the connection to PPR, we develop
a proveably-correct approximate inference scheme, and an associated
proveably-correct approximate grounding scheme: specifically, we show
that it is possible to prove a query, or to build a graph which
contains the information necessary for weight-learning, in time
$O(\frac{1}{\alpha\epsilon})$, where $\alpha$ is a reset parameter
associated with the bias towards short derivations, and $\epsilon$ is
the worst-case approximation error across all intermediate stages of
the proof.  This means that both inference and learning can be
approximated in time \emph{independent of the size of the underlying
  database}---a surprising and important result.

The ability to locally ground queries has another important
consequence: it is possible to decompose the problem of
weight-learning to a number of moderate-size subtasks (in fact, tasks
of size $O(\frac{1}{\alpha\epsilon})$ or less) which are weakly
coupled.  Based on this we outline a parallelization scheme, which in
our initial implementation provides a order-of-magnitude speedup in
learning time.

Below, we will first introduce our formalism, and then describe our
weight-learning algorithm.  We will then present experimental results
on a prototypical inference task, and compare the scalability of our
method to Markov logic networks.  We finally discuss related work and
conclude.

\section{\underline{Pro}gramming with \underline{P}ersonalized
  \underline{P}age\underline{R}ank (PROPPR)}

\subsection{LOGIC PROGRAM INFERENCE AS GRAPH SEARCH}

We will now describe our ``locally groundable'' first-order
probabilistic language, which we call ProPPR.
Inference for ProPPR is based on
a personalized PageRank process over the proof constructed by Prolog's
Selective Linear Definite (SLD) theorem-prover.  To define the
semantics we will use notation from logic programming \cite{Lloyd}.
Let $LP$ be a program which contains a set of definite clauses
$c_1,\ldots,c_n$, and consider a conjunctive query $Q$ over the
predicates appearing in $LP$.  A traditional Prolog interpreter can be
viewed as having the following actions.  First, construct a ``root
vertex'' $v_0$ which is a pair $(Q,Q)$ and add it to an
otherwise-empty graph $G'_{Q,LP}$. (For brevity, we will use drop the
subscripts of $G'$ where possible.)  Then recursively add to $G'$ new
vertices and edges as follows: if $u$ is a vertex of the form $(Q,
(R_1,\ldots,R_k))$, and $c$ is a clause in $LP$ of the form \( R'
\leftarrow S'_1,\ldots,S'_\ell \), and $R_1$ and $R'$ have a most
general unifier $\theta=mgu(R_1,R')$, then add to $G'$ a new edge
$\edgeuv$ where \( v = (Q\theta,
(S'_1,\ldots,S'_\ell,R2,\ldots,R_k)\theta) \).  Let us call $Q\theta$
the \trm{transformed query} and
$(S'_1,\ldots,S'_\ell,R2,\ldots,R_k)\theta$ the \trm{associated
  subgoal list}.  If a subgoal list is empty, we will denote it by
$\Box$.

\begin{figure*}
\centerline{\includegraphics[scale=0.35]{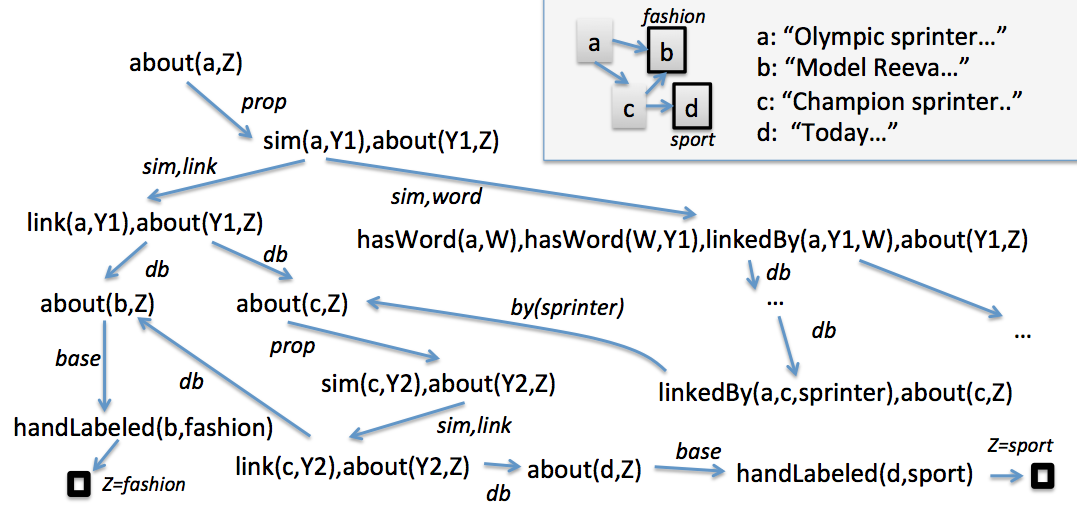}}
\caption{A partial proof graph for the query \textit{about(a,Z)}.  The
  upper right shows the link structure between documents $a,b,c$, and
  $d$, and some of the words in the documents.  Restart links are not
  shown.}
\label{fig:proof}
\end{figure*}

$G'$ is often large or infinite so it is not constructed
explicitly. Instead Prolog performs a depth-first search on $G'$ to
find the first \trm{solution vertex} $v$---i.e., a vertex with an
empty subgoal list---and if one is found, returns the transformed
query from $v$ as an answer to $Q$.  Table~\ref{tab:proppr} and
Figure~\ref{fig:proof} show a simple Prolog program and a proof graph
for it.\footnote{The annotations after the hashmarks and the edge
  labels in the proof graph will be described below. For conciseness,
  only $R_1,\ldots,R_k$ is shown in each node
  $u=(Q,(R_1,\ldots,R_k))$.}  Given the query $Q=\textit{about(a,Z)}$,
Prolog's depth-first search would return
$Q=\textit{about(a,fashion)}$.

Note that in this proof formulation, the nodes are \emph{conjunctions}
of literals, and the structure is, in general, a digraph (rather than
a tree). Also note that the proof is encoded as a graph, not a hypergraph,
even if the predicates in the LP are not binary: the edges represent a
step in the proof that reduces one conjunction to another, not a
binary relation between entities.

\begin{table}
\caption{A simple program in ProPPR.  See text for explanation.} \label{tab:proppr}
%\begin{small}
~\\
\begin{tabular}{|ll|}
\hline
about(X,Z) :- handLabeled(X,Z)    &\# base. \\
about(X,Z) :- sim(X,Y),about(Y,Z)   &\# prop. \\
sim(X,Y) :- links(X,Y)                      &\# sim,link. \\
sim(X,Y) :-                                 & \\
~~~~hasWord(X,W),hasWord(Y,W),              & \\
~~~~linkedBy(X,Y,W)                         & \# sim,word.\\
linkedBy(X,Y,W) :- true                     & \# by(W). \\
\hline
\end{tabular}
%\end{small}
\end{table}

\subsection{FROM STOCHASTIC LOGIC PROGRAMS TO PROPPR}

In \trm{stochastic logic programs} (SLPs)
\cite{DBLP:journals/ml/Cussens01}, one defines a randomized procedure
for traversing the graph $G'$ which thus defines a probability
distribution over vertices $v$, and hence (by selecting only solution
vertices) a distribution over transformed queries (i.e. answers)
$Q\theta$.  The randomized procedure thus produces a distribution over
possible answers, which can be tuned by learning to upweight desired
(correct) answers and downweight others.

\wc{add more discussion of SLPs - experiments, comparisons to other LPs}

In past work, the randomized traversal of $G'$ was defined by a
probabilistic choice, at each node, of which clause to apply, based on
a weight for each clause.  We propose two extensions.  First, we will
introduce a new way of computing clause weights, which allows for a
potentially richer parameterization of the traversal process.  We will
associate with each edge $\edgeuv$ in the graph a \trm{feature vector}
$\phi_{\edgeuv}$.  This edge is produced indirectly, by associating
with every clause $c\in{}LP$ a function $\Phi_c(\theta)$, which
produces the $\phi$ associated with an application of $c$ using mgu
$\theta$.  This feature vector is computed during theorem-proving, and
used to annotate the edge $\edgeuv$ in $G'$ created by applying $c$
with mgu $\theta$. Finally, an edge $\edgeuv$ will be traversed with
probability \( \Pr(v|u) \propto f(\weightvec,\phi_\edgeuv) \) where
$\weightvec$ is a parameter vector and where $f(\weightvec,\phi)$ is a
weighting function---e.g., $f(\weightvec,\phi) = \exp(
\weightvec_i\cdot \phi)$.  This weighting function now determines the
probability of a transition, in theorem-proving, from $u$ to $v$:
specifically, $\Pr_\weightvec(v|u) \propto
f(\weightvec,\phi_\edgeuv)$.  Weights in $\weightvec$ default to 1.0,
and learning consists of tuning these.

The second and more fundamental extension is to add edges in $G'$ from
every solution vertex to itself, and also add an edge from every
vertex to the start vertex $v_0$. We will call this augmented graph
$G_{Q,LP}$ below (or just $G$ if the subscripts are clear from
context).  These links make SLP's graph traversal a \trm{personalized
  PageRank} (PPR) procedure\uncite{pagerank,SoumenWWW2007}, sometimes
known as \trm{random-walk-with-restart}
\cite{DBLP:conf/icdm/TongFP06}.  These links are annotated by another
feature vector function $\Phi_{restart}(R)$, which is a applied of the
leftmost literal of the subgoal list for $u$ to annotate the edge
$\edge{u}{v_0}$.

These links back to the start vertex bias will the traversal of the
proof graph to upweight the results of \emph{short proofs}. To see
this, note that if the restart probability $P(v_0|u)=\alpha$ for every
node $u$, then the probability of reaching any node at depth $d$ is
bounded by $(1-\alpha)^d$.

To summarize, if $u$ is a node of the search graph,
$u=(Q\theta,(R_1,\ldots,R_k))$, then the transitions from $u$, and
their respective probabilities, are defined as follows, where $Z$ is
an appropriate normalizing constant:
\begin{itemize}
\item If \( v = (Q\theta\sigma,
  (S'_1,\ldots,S'_\ell,R2,\ldots,R_k)\theta\sigma) \) is a state
  derived by applying the clause $c$ (with mgu $\sigma$), then
\[ \Pr_\weightvec(v|u)=\frac{1}{Z} f(\weightvec,\Phi_c(\theta\circ\sigma))
\]
\item If \( v = v_0 = (Q,Q) \) is the initial state in $G$, then
\[ \Pr_\weightvec(v|u)=\frac{1}{Z} f(\weightvec,\Phi_{restart}(R_1\theta))
\]
\item If $v$ is any other node, then $\Pr(v|u)=0$.
\end{itemize}

Finally we must specify the functions $\Phi_c$ and $\Phi_{restart}$.
For clauses in $LP$, the feature-vector producing function
$\Phi_c(\theta)$ for a clause is specified by annotating $c$ as
follows: every clause $c=(R\leftarrow S_1,\ldots,S_k)$ can be
annotated with an additional conjunction of ``feature literals''
$F_1,\ldots,F_\ell$, which are written at the end of the clause after
the special marker ``\#''.  The function $\Phi_c(\theta)$ then returns
a vector $\phi=\{F_1\theta,\ldots,F_\ell\theta\}$, where every
$F_i\theta$ must be ground.

The requirement\footnote{The requirement that the feature literals
  returned by $\phi_c(\theta)$ must be ground in $\theta$ is not
  strictly necessary for correctness.  However, in developing ProPPR
  programs we noted than non-ground features were usually not what the
  programmer intended.} that edge features $F_i\theta$ are ground is
the reason for introducing the apparently unnecessary predicate
\textit{linkedBy(X,Y,W)} into the program of Table~\ref{tab:proppr}:
adding the feature literal \textit{by(W)} to the second clause for
\textit{sim} would result in a non-ground feature \textit{by(W)},
since $W$ is a free variable when $\Phi_c$ is called.  Notice also
that the weight on the \textit{by(W)} features are meaningful, even
though there is only one clause in the definition of
\textit{linkedBy}, as the weight for applying this clause competes
with the weight assigned to the restart edges.

It would be cumbersome to annotate every database fact, and difficult
to learn weights for so many features. Thus, if $c$ is the unit clause
that corresponds to a database fact, then $\Phi_c(\theta)$ returns a
default value $\phi=\{\textit{db}\}$, where \textit{db} is a special
feature indicating that a database predicate was used.\footnote{If a
  non-database clause $c$ has no annotation, then the default vector
  is $\phi=\{\textit{id(c)}\}$, where \textit{c} is an identifier for
  the clause $c$.}

The function $\Phi_{restart}(R)$ depends on the functor and arity of
$R$.  If $R$ is defined by clauses in $LP$, then $\Phi_{restart}(R)$
returns a unit vector $\phi=\{\textit{defRestart}\}$.  If $R$ is a
database predicate (e.g., \textit{hasWord(doc1,W)}) then we follow a
slightly different procedure, which is designed to ensure that the
restart link has a reasonably large weight even with unit feature
weights: we compute $n$, the number of possible bindings for $R$, and
set $\phi[\textit{defRestart}] = n\cdot{}\frac{\alpha}{1-\alpha}$,
where $\alpha$ is a global parameter.  This means that with unit
weights, after normalization, the probability of following the restart
link will be $\alpha$.

Putting this all together with the standard iterative approach to
computing personalized PageRank over a graph \cite{pagerank}, we
arrive at the following inference algorithm for answering a query $Q$,
using a weight vector $\weightvec$.  Below, we let $N_{v_0}(u)$ denote
the \trm{neighbors} of $u$---i.e., the set of nodes $v$ where
$\Pr(v|u)>0$ (including the restart node $v=v_0$).  We also let
\textbf{W} be a matrix such that
$\textbf{W}[u,v]=\Pr_\weightvec(v|u)$, and in our discussion, we use
$\textbf{ppr}(v_0)$ to denote the personalized PageRank vector for
$v_0$.
\begin{enumerate}
\item Let $v_0=(Q,Q)$ be the start node of the search graph.
  Let $G$ be a graph containing just $v_0$. Let $\textbf{v}^0 = \{ v_0 \}$.
\item For $t=1,\ldots,T$ (i.e., until convergence):
  \begin{enumerate}
    \item Let $\textbf{v}^{t}$ be an all-zeros vector.
    \item For each $u$ with non-zero weight in $\textbf{v}^{t-1}$, and
      each $v\in{}N_{u+0}(u)$, add $(u,v,\phi_{\edgeuv})$ to $G$ with weight
      $\Pr_\weightvec(v|u)$, and set 
      $\textbf{v}^{t} = \textbf{W}\cdot\textbf{v}^{t-1}$
  \end{enumerate}
\item At this point $\textbf{v}^T \approx \textbf{ppr}(v_0)$.  Let $S$
  be the set of nodes $(Q\theta,\Box)$ that have empty subgoal lists
  and non-zero weight in $\textbf{v}^T$, and let $Z=\sum_{u\in{}S}
  \textbf{v}^T[u]$.  The final probability for the literal $L=Q\theta$
  is found by extracting these solution nodes $S$, and renormalizing:
  \[ 
     \Pr_\weightvec(L) \equiv \frac{1}{Z} \textbf{v}^T[(L,\Box)]
  \]
\end{enumerate}
For example, given the query $Q=\textit{about(a,Z)}$ and the program
of Table~\ref{tab:proppr}, this procedure would give assign a non-zero
probability to the literals $\textit{about(a,sport)}$ and
$\textit{about(a,fashion)}$, concurrently building the graph of
Figure~\ref{fig:proof}.

\wc{discussion: not real useful for ground queries.}

\subsection{LOCALLY GROUNDING A QUERY}

\begin{table*}
\caption{The PageRank-Nibble-Prove algorithm for inference in ProPPR.
  $\alpha'$ is a lower-bound on $\Pr(v_0|u)$ for any node $u$ to be
  added to the graph $\hat{G}$, and $\epsilon$ is the desired degree
  of approximation.}
\label{tab:pr-nibble}
\begin{center}
\begin{minipage}[t]{\textwidth}
\begin{tabbing}1234\=1234\=1234\=1234\=1234\=1234\=1234\=1234\=1234\=\kill
define PageRank-Nibble-Prove($Q$):\\
\>let $\textbf{v} = $PageRank-Nibble$((Q,Q),\alpha',\epsilon)$\\
\>let $S=\{ u:\textbf{p}[u]>u \mbox{~and~} u=(Q\theta,\Box) \}$\\
\>let $Z=\sum_{u\in S} \textbf{p}[u]$\\
\>define $\Pr_\weightvec(L) \equiv \frac{1}{Z} \textbf{v}[(L,\Box)]$\\
end\\
~\\
define PageRank-Nibble($v_0,\alpha',\epsilon$):\\
\>let $\textbf{p} = \textbf{r} = \textbf{0}$, let $\textbf{r}[v_0] = 1$, and let $\hat{G}=\emptyset$\\
\>while $\exists u: \textbf{r}(u)/|N(u)| > \epsilon$ do: push($u$)\\
\> return $\textbf{p}$\\
end\\
\end{tabbing}\end{minipage}~~~~~~~\begin{minipage}[t]{\textwidth}
\begin{tabbing}1234\=1234\=1234\=1234\=1234\=1234\=1234\=1234\=1234\=\kill
define push($u$):\\
\> \textit{comment: this modifies \textbf{p}, \textbf{r}, and $\hat{G}$}\\
\> $\textbf{p}[u] = \textbf{p}[u] + \alpha'\cdot\textbf{r}[u]$\\
\> $\textbf{r}[u] = \textbf{r}[u] \cdot (1 - \alpha')$\\
\> for $v\in{}N(u)$:\\
\> \> add the edge $(u,v,\phi_{\edgeuv})$ to $\hat{G}$\\
\> \> if $v=v_0$ then $\textbf{r}[v] = \textbf{r}[v] + \Pr(v|u)\textbf{r}[u]$\\
\> \> else $\textbf{r}[v] = \textbf{r}[v] + {(\Pr(v|u)-\alpha')}\textbf{r}[u]$\\
\> endfor\\
end
\end{tabbing}\end{minipage}
\end{center}
\end{table*}

Note that this procedure both performs inference (by computing a
distribution over literals $Q\theta$) and ``grounds'' the query, by
constructing a graph $G$.  ProPPR inference for this query can be
re-done efficiently, by running an ordinary PPR process on $G$. This
is useful for faster weight learning. Unfortunately, the grounding $G$
can be very large: it need not include the entire database, but if $T$
is the number of iterations until convergence for the sample program
of Table~\ref{tab:proppr} on the query $Q=about(d,Y)$, $G$ will
include a node for every page within $T$ hyperlinks of $d$.

To construct a more compact local grounding graph $G$, we adapt an
approximate personalized PageRank method called PageRank-Nibble
\cite{DBLP:journals/im/AndersenCL08}.  This method has been used for
the problem of \trm{local partitioning}: in local partitioning, the
goal is to find a small, low-conductance\footnote{For small subgraphs
  $G_S$, \trm{conductance} of $G_S$ is the ratio of the weight of all
  edges exiting $G_S$ to the weight of all edges incident on a node in
  $G_S$.}  component of a large graph $G$ that contains a given node
$v$.

The PageRank-Nibble-Prove algorithm is shown in
Table~\ref{tab:pr-nibble}.  It maintains two vectors: $\textbf{p}$, an
approximation to the personalized PageRank vector associated with node
$v_0$, and $\textbf{r}$, a vector of ``residual errors'' in
$\textbf{p}$.  Initially, $\textbf{p}=\emptyset$ and
$\textbf{r}=\{v_0\}$.  The algorithm repeatedly picks a node $u$ with
a large residual error $\textbf{r}[u]$, and reduces this error by
distributing a fraction $\alpha'$ of it to $\textbf{p}[u]$, and the
remaining fraction back to $\textbf{r}[u]$ and
$\textbf{r}[v_1],\ldots,\textbf{r}[v_n]$, where the $v_i$'s are the
neighbors of $u$.  The order in which nodes $u$ are picked does not
matter for the analysis (in our implementation, we follow Prolog's
usual depth-first search as much as possible.)  Relative to
PageRank-Nibble, the main differences are the the use of a lower-bound
on $\alpha$ rather than a fixed restart weight and the construction of
the graph $\hat{G}$.

Following the proof technique of Andersen et al, it can be shown that
after each push, $\textbf{p}+\textbf{r}=\textbf{ppr}(v_0)$.  It is
also clear than when PageRank-Nibble terminates, then for any $u$, the
error $\textbf{ppr}(v_0)[u] - \textbf{p}[u]$ is bounded by
$\epsilon{}N(u)$: hence, in any graph where $N(u)$ is bounded, a good
approximation can be obtained.  It can also be shown
\cite{DBLP:journals/im/AndersenCL08} that the subgraph $\hat{G}$ (of
the full proof space) is in some sense a ``useful'' subset: for an
appropriate setting of $\epsilon$, if there is a low-conductance
subgraph $G_*$ of the full graph that contains $v_0$, then $G_*$ will
be contained in $\hat{G}$: thus if there is a subgraph $G_*$
containing $v_0$ that approximates the full graph well,
PageRank-Nibble will find (a supergraph of) $G_*$.

Finally, we have the following efficiency bound:

\begin{theorem}[Andersen,Chung,Lang]
Let $u_i$ be the $i$-th node pushed by PageRank-Nibble-Prove.  Then
$\sum_{i} |N(u_i)| < \frac{1}{\alpha'\epsilon}$.
\end{theorem}

This can be proved by noting that initially $|\textbf{r}|_1=1$, and
also that $|\textbf{r}|_1$ decreases by at least
$\alpha'\epsilon|N(u_i)|$ on the $i$-th push.  As a direct
consequence we have the following:

\begin{corollary}
The number of edges in the graph $\hat{G}$ produced by PageRank-Nibble-Prove
is no more than $\frac{1}{\alpha'\epsilon}$.
\end{corollary}

Importantly, the bound holds \emph{independent of the size of the full
  database of facts}.  The bound also holds regardless of the size or
loopiness of the full proof graph, so this inference procedure will
work for recursive logic programs.

To summarize, we have outlined an efficient approximate proof
procedure, which is closely related to personalized PageRank.  As a
side-effect of inference for a query $Q$, this procedure will create a
ground graph $\hat{G}_Q$ on which personalized PageRank can be run directly,
without any (relatively expensive) manipulation of first-order
theorem-proving constructs such as clauses or logical variables.  As
we will see, this ``locally grounded'' graph will be very useful in
learning weights $\weightvec$ to assign to the features of a ProPPR
program.

\begin{table*}
\caption{Some more sample ProPPR programs.  $LP=\{c_1,c_2\}$ is a
  bag-of-words classifier (see text).  $LP=\{c_1,c_2,c_3,c_4\}$ is a
  recursive label-propagation scheme, in which predicted labels for
  one document are assigned to similar documents, with similarity
  being an (untrained) cosine distance-like
  measure. $LP=\{c_1,c_2,c_5,c_6\}$ is a sequential classifier for
  document sequences.}
\label{tab:sample}

%\begin{tabular}[ll]
\begin{center}
\begin{minipage}[t]{0.4\textwidth}
\begin{tabbing}
$c_1$: \=pred\=ictedClass(Doc,Y) :- \\
\> \>possibleClass(Y),\\
\> \>hasWord(Doc,W),\\
\> \>related(W,Y) \# c1.\\
$c_2$: related(W,Y) :- true\\
\>\# relatedFeature(W,Y)\\
~\\
~\\
\emph{Database predicates}:\\
\emph{hasWord(D,W): doc $D$ contains word $W$}\\
\emph{inDoc(W,D): doc $D$ contains word $W$}\\
\emph{previous(D1,D2): doc $D2$ precedes $D1$}\\
\emph{possibleClass(Y): $Y$ is a class label}\\
\end{tabbing}
\end{minipage}~~~~~~\begin{minipage}[t]{0.4\textwidth}
\begin{tabbing}
$c_3$: \=pre\=dictedClass(Doc,Y) :- \\
\> \>similar(Doc,OtherDoc),\\
\> \>predictedClass(OtherDoc,Y) \# c3.\\
$c_4:$ similar(Doc1,Doc2) :- \\
\> \>hasWord(Doc1,W),\\
\> \>inDoc(W,Doc2) \# c4.\\
~\\
$c_5:$ predictedClass(Doc,Y) :- \\
\> \>previous(Doc,OtherDoc),\\
\> \>predictedClass(OtherDoc,OtherY),\\
\> \>transition(OtherY,Y) \# c5.\\
$c_6$: transition(Y1,Y2) :- true \\
\> \# transitionFeature(Y1,Y2) \\
\end{tabbing}
\end{minipage}
\end{center}
%\end{tabular}
\end{table*}

As an illustration of the sorts of ProPPR programs that are possible,
some small sample programs are shown in Figure~\ref{tab:sample}.
Clauses $c_1$ and $c_2$ are, together, a bag-of-words classifier: each
proof of \emph{predictedClass(D,Y)} adds some evidence for $D$ having
class $Y$, with the weight of this evidence depending on the weight
given to $c_2$'s use in establishing \emph{related(w,y)}, where $w$
and $y$ are a specific word in $D$ and $y$ is a possible class
label. In turn, $c_2$'s weight depends on the weight assigned to the
$r(w,y)$ feature by $\weightvec$, relative to the weight of the
restart link.\footnote{The existence of the restart link thus has
  another important role in this program, as it avoids a sort of
  ``label bias problem'' \uncite{LaffertyML2001} in which local
  decisions are difficult to adjust.}  Adding $c_3$ and $c_4$ to this
program implements label propagation, and adding $c_5$ and $c_6$
implements a sequential classifier.

In spite of its efficient inference procedure, and its limitation to
only definite clauses, ProPPR appears to have much of the expressive
power of MLNs \cite{DBLP:series/synthesis/2009Domingos}, in that many
useful heuristics can apparently be encoded.

\subsection{LEARNING FOR PROPPR}

As noted above, inference for a query $Q$ in ProPPR is based on a
personalized PageRank process over the graph associated with the SLD
proof of a query goal $G$.  More specifically, the edges $\edgeuv$ of
the graph $G$ are annotated with feature vectors $\phi_{\edgeuv}$, and
from these feature vectors, weights are computed using a parameter
vector $\weightvec$, and finally normalized to form a probability
distribution over the neighbors of $u$.  The ``grounded'' version of
inference is thus a personalized PageRank process over a graph with
feature-vector annotated edges.

In prior work, Backstrom and Leskovic \cite{BackstromKDD2006} outlined
a family of supervised learning procedures for this sort of annotated
graph.  In the simpler case of their learning procedure, an example is
a triple $(v_0,u,y)$ where $v_0$ is a query node, $u$ is a node in in
the personalized PageRank vector $\textbf{p}_{v_0}$ for $v_0$, $y$ is
a target value, and a loss $\ell(v_0,u,y)$ is incurred if
$\textbf{p}_{v_0}[u] \not= y$.  In the more complex case of ``learning
to rank'', an example is a triple $(v_0,u_+,u_-)$ where $v_0$ is a
query node, $u_+$ and $u_-$ are nodes in in the personalized PageRank
vector $\textbf{p}_{v_0}$ for $v_0$, and a loss is incurred unless
$\textbf{p}_{v_0}[u_+] \geq \textbf{p}_{v_0}[u_-]$.  The core of
Backstrom and Leskovic's result is a method for computing the gradient
of the loss on an example, given a differentiable feature-weighting
function $f(\weightvec,\phi)$ and a differentiable loss function
$\ell$.  The gradient computation is broadly similar to the
power-iteration method for computation of the personalized PageRank
vector for $v_0$. Given the gradient, a number of optimization methods
can be used to compute a local optimum.
 
We adopt this learning for ProPPR, with some modifications.  The
training data $D$ is a set of triples $\{
(Q^1,P^1,N^1),\ldots,(Q^m,P^m,N^m)\}$ where each $Q^k$ is a query,
$P^k=\langle Q\theta_+^1,\ldots,Q\theta_+^I\rangle$ is a list of
correct answers, and $N^k$ is a list $\langle
Q\theta_-^1,\ldots,Q\theta_-^J\rangle$ incorrect answers.  Each such
triple is then locally grounded using the PageRank-Nibble-Prove method
and used to produce a set $I*J$ of ``learning-to-order'' triples of
the form $(v^k_0,u_+^{k,i},u_-^{k,j})$ where $v_0^k$ corresponds to
$Q^k$, and the $u$'s are the nodes in $\hat{G}_{Q^K}$ that correspond
to the (in)correct answers for $Q^K$.  We use a squared loss on the
difference of scores $h=\textbf{p}_{v_0}[u_+] -
\textbf{p}_{v_0}[u_-]$, i.e.,
\[
 \ell(v_0,u_+,u_-) \equiv 
  \left\{
    \begin{array}{cc} h^2 &  \mbox{if $h<0$} \\
      0 &  else \\
    \end{array}
   \right.
\] 
and $L_2$ regularization of the parameter weights.  Hence the final
function to be optimized is
\[
\sum_{k} \sum_{i,j} \ell(v^k_0,u_+^{k,i},u_-^{k,j}) + \mu||\textbf{w}||^2_2
\]
To optimize this loss, we use stochastic gradient descent (SGD),
rather than the quasi-Newton method of Backstrom and Leskovic.
Weights are initialized to $1.0+\delta$, where $\delta$ is randomly
drawn from $[0,0.01]$.  We set the learning rate $\beta$ of SGD to be
\(
\beta = \frac{\eta}{epoch^2}
\)
where $epoch$ is the current epoch in SGD, and $\eta$, the initial learning rate,
defaults to 1.0.
%WW2: there's no definition of learning rate update, so I added here.

We implemented SGD because it is fast and has
been adapted to parallel learning tasks
\cite{zinkevich2010parallelized,niu2011hogwild}.  Local grounding
means that learning for ProPPR is quite well-suited to
parallelization.  The step of locally grounding each $Q_i$ is
``embarassingly'' parallel, as every grounding can be done
independently.  To parallelize the weight-learning stage, we use
multiple threads, each of which computes the gradient over a single
grounding $\hat{G}_{Q^k}$, and all of which accesses a single shared
parameter vector $\weightvec$.  Although the shared parameter vector
is a potential bottleneck \cite{zinkevich2009slow}, it is not a severe
one, as the gradient computation dominates the learning
cost.\footnote{This is not the case when learning a linear classifier,
  where gradient computations are much cheaper.}

\section{EXPERIMENTS}
%WW: in general, my feeling about the experiment section is that most 
% of the important bits are in the figures and tables, but 
% maybe we want to explain/discuss more about the results in the text?
\subsection{A SAMPLE TASK}

\begin{table*}
\caption{ProPPR program used for entity resolution.} \label{tab:erprog}
\begin{small}
\begin{tabular}{|ll|}
\hline
samebib(BC1,BC2) :- author(BC1,A1),sameauthor(A1,A2),authorinverse(A2,BC2) & \# author.\\
samebib(BC1,BC2) :- title(BC1,A1),sametitle(A1,A2),titleinverse(A2,BC2) & \# title.\\
samebib(BC1,BC2) :- venue(BC1,A1),samevenue(A1,A2),venueinverse(A2,BC2) & \# venue.\\
samebib(BC1,BC2) :- samebib(BC1,BC3),samebib(BC3,BC2) & \# tcbib.\\ %\hline
% & \\
sameauthor(A1,A2) :- haswordauthor(A1,W),haswordauthorinverse(W,A2),keyauthorword(W) & \# authorword.\\
sameauthor(A1,A2) :- sameauthor(A1,A3),sameauthor(A3,A2) & \# tcauthor.\\ %\hline
%& \\
sametitle(A1,A2) :- haswordtitle(A1,W),haswordtitleinverse(W,A2),keytitleword(W) & \# titleword.\\
sametitle(A1,A2) :- sametitle(A1,A3),sametitle(A3,A2) & \# tctitle.\\ %\hline
%& \\
samevenue(A1,A2) :- haswordvenue(A1,W),haswordvenueinverse(W,A2),keyvenueword(W) & \# venueword.\\
samevenue(A1,A2) :- samevenue(A1,A3),samevenue(A3,A2) & \# tcvenue.\\ %\hline
%& \\
keyauthorword(W) :- true & \#  authorWord(W). \\
keytitleword(W) :- true & \#  titleWord(W). \\
keyvenueword(W) :- true & \#  venueWord(W). \\
\hline
\end{tabular}
\end{small}
\end{table*}

To evaluate this method, we use data from an entity resolution task
previously studied as a test case for MLNs \cite{singla2006entity}.
The program we use in the experiments is shown in
Table~\ref{tab:erprog}: it is approximately the same as the MLN(B+T)
approach from Singla and Domingos.\footnote{The principle difference
  is that we do not include tests on the absence of words in a field in
  our clauses.}  To evaluate accuracy, we use the CORA dataset, a
collection of 1295 bibliography citations that refer to 132 distinct
papers. Throughout the experiments, we set the regularization coefficient
$\mu$ to $0.001$, the total number of epochs to 5, and learning rate parameter $\eta$
to 1. A standard log loss function was used in 
our objective function.
%WW2: adding the parameter values above.
\wc{where we get extra data from, to explore scalability against DB
  size}

\subsection{RESULTS}

% untrained ProPPR on 52 problems
%epsilon	    map	       		      time(s)
%0.001	    0.0377358490566	      7			      
%0.0001	    0.297480954626	      28
%0.00005	    0.39697867474	      39
%0.00002	    0.532478145586	      75
%0.00001	    0.539898788814	      116
%0.000005    0.542811175823	      216
%ppr	    0.540505233578	      819

\begin{table}
\caption{Performance of the approximate PageRank-Nibble-Prove method,
  compared to the grounding by running personalized PageRank to
  convergence.  In all cases $\alpha'=0.1$.} \label{tab:dprVsPPR}
\begin{center}
\begin{tabular}{lrr}
$\epsilon$ & MAP & Time(sec) \\
\hline
0.0001	    & 0.30	      & 28 \\
0.00005	    & 0.40	      & 39 \\
0.00002	    & 0.53	      & 75 \\
0.00001	    & 0.54	      & 116 \\
0.000005    & 0.54	      & 216 \\
\hline
power iteration
            & 0.54	      & 819 \\
\end{tabular}
\end{center}
\end{table}

We first consider the cost of the PageRank-Nibble-Prove
inference/grounding technique.  Table~\ref{tab:dprVsPPR} shows the
time required for inference (with uniform weights) for a set of 52
randomly chosen entity-resolution tasks from the CORA dataset, using a
Python implemention of the theorem-prover.  We report the time in
seconds for all 52 tasks, as well as the mean average precision (MAP)
of the scoring for each query.  It is clear that PageRank-Nibble-Prove
offers a substantial speedup on these problems with little loss in
accuracy: on these problems, the same level of accuracy is achieved in
less than a tenth of the time.

\begin{figure}
\centerline{\includegraphics[scale=0.62]{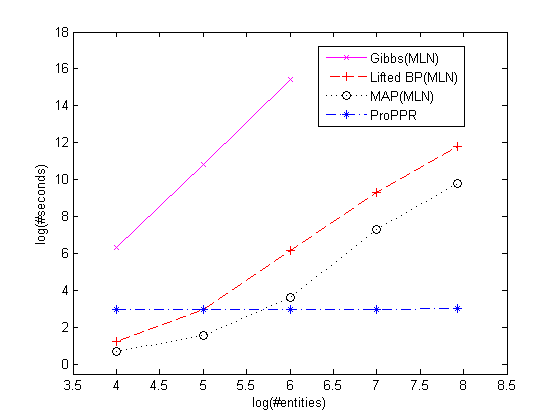}}
\caption{Run-time for inference in ProPPR (with a single thread) as a
  function of the number of entities in the database.  The base of the
  log is 2.}
\label{fig:time}
\end{figure}

While the speedup in inference time is desirable, the more important
advantages of the local grounding approach are that (1) grounding
time, and hence inference, need not grow with the database size and
(2) learning can be performed in parallel, by using multiple threads
for parallel computations of gradients in SGD.  Figure~\ref{fig:time}
illustrates the first of these points: the scalability of the
PageRank-Nibble-Prove method as database size increases.  For
comparison, we also show the inference time for MLNs with three
well-published inference methods: Gibbs refers to Gibbs sampling, and
Lifted BP is the lifted belief propagation method.  We also compare
with the maximum a posteriori (MAP) inference approach, which does not
return probabilistic estimates of the specified queries.  In each case
the performance task is inference over 16 test queries. 

Note that ProPPR's runtime is constant, independent of the database
size: it takes essentially the same time for $2^8=256$ entities as for
$2^4=16$.  In contrast, lifted belief propagation is around 1000 times
slower on the larger database.

%trained:
%Citation prediction    0.79968 (0.00760)
%Author prediction     0.83986 (0.02134) 
%Venue prediction      0.86850 (0.01135)
%Title prediction        0.90074 (0.01663)	
%
%untrained:
%Citation prediction   0.67980 (0.00938)
%Author prediction     0.83733 (0.02147)
%Venue prediction     0.85998 (0.01165)
%Title prediction        0.90783 (0.01612)

\begin{figure}[h]
%\centerline{\fbox{\includegraphics[angle=-90,width=0.56\textwidth]{speedup.png}}} % was: scale=0.3
\includegraphics[angle=0,width=0.53\textwidth]{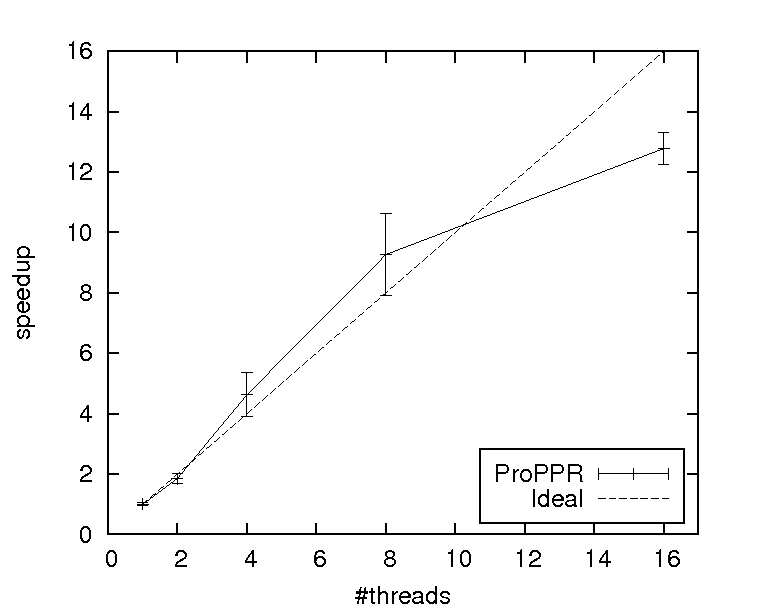}
\caption{Performance of the parallel SGD method.  The $x$ axis is the
  number of threads on a multicore machine, and the $y$ axis is the
  speedup factor over a single-threaded implementation.}
\label{fig:threads}
\end{figure}

\begin{table}
\caption{AUC results on CORA citation-matching.} \label{tab:auc}
~\\
\begin{tabular}{lrrrr}
                      
& Cites & Authors & Venues & Titles \\ \hline
MLN(Fig~\ref{tab:proppr})   & 0.513  & 0.532  & 0.602 & 0.544\\
MLN(S\&D)                   & 0.520  & 0.573  & 0.627 & 0.629\\
ProPPR(\textbf{w}=1)        & 0.680  & 0.836  & 0.860 & \textbf{0.908} \\
ProPPR                      & \textbf{0.800}  & \textbf{0.840}  & \textbf{0.869} & 0.900 \\
%WW: is this enough, or should we add other results?
%MLN(B+T)           & 0.950     & 0.994  & 0.745 & $-$ \\
\hline
\end{tabular}
\end{table}

Figure~\ref{fig:threads} explores the speedup in learning (from
grounded examples) due to multi-threading.  The weight-learning is
using a Java implementation of the algorithm which runs over ground
graphs.  The full CORA dataset was used in this experiment.  As can be
seen, the speedup that is obtained is nearly optimal, even with 16
threads running concurrently.

We finally consider the effectiveness of weight learning.  
We train on the first four sections of the CORA dataset,
and report results on the fifth.  Following
Singla and Domingos \cite{singla2006entity} we report performance as
area under the ROC curve (AUC).  Table~\ref{tab:auc} shows AUC on the
test set used by Singla and Domingos for several methods.  The line
for MLN(Fig~\ref{tab:proppr}) shows results obtained by an MLN version
of the program of Figure~\ref{tab:proppr}.  The line MLN(S\&D) shows
analogous results for the best-performing MLN from
\cite{singla2006entity}.  Compared to these methods, ProPPR does quite
well even before training (with unit feature weights, \textbf{w}=1);
the improvement here is likely due to the ProPPR's bias towards short
proofs, and the tendency of the PPR method to put more weight on
shared words that are rare (and hence have lower fanout in the graph
walk.)  Training ProPPR improves performance on three of the four
tasks, and gives the most improvement on citation-matching, the most
complex task.

The results in Table~\ref{tab:auc} all use the same data and
evaluation procedure, and the MLNs were trained with the
state-of-the-art Alchemy system using the recommended commands for
this data (which is distributed with
Alchemy\footnote{http://alchemy.cs.washington.edu}).  However, we
should note that the MLN results reproduced here are not identical to
previous-reported ones \cite{singla2006entity}. Singla and Domingos
used a number of complex heuristics that are difficult to
reproduce---e.g., one of these was combining MLNs with a heuristic,
TFIDF-based matching procedure based on canopies
\cite{mccallum00efficient}.  While the trained ProPPR model
outperformed the reproduced MLN model in all prediction tasks, it
outperforms the reported results from Singla and Domingos only on
\textit{venue}, and does less well than the reported results on
\textit{citation} and \textit{author}\wc{, which have AUC of 0.95 and
  0.99 respectively using the MLN program most similar to the ProPPR
  program used here.}\footnote{Performance on \textit{title} matching
  is not reported by Singla and Domingos.}.

\section{RELATED WORK}

Although we have chosen here to compare experimentally to MLNs
\cite{RichardsonMLJ2006,singla2006entity}, ProPPR represents a rather
different philosophy toward language design: rather than beginning
with a highly-expressive but intractible logical core, we begin with a
limited logical inference scheme and add to it a minimal set of
extensions that allow probabilistic reasoning, while maintaining
stable, efficient inference and learning.  While ProPPR is less
expressive than MLNs (for instance, it is limited to definite clause
theories) it is also much more efficient.  This philosophy is similar
to that illustrated by probabilistic similarity logic (PSL)
\cite{brocheler2012probabilistic}; however, unlike ProPPR, PSL does
not include a ``local'' grounding procedure, which leads to small
inference problems, even for large databases.

Technically, ProPPR is most similar to stochastic logic programs
(SLPs) \cite{DBLP:journals/ml/Cussens01}.  The key innovation is the
integration of a restart into the random-walk process, which, as we
have seen, leads to very different computational properties.

There has been some prior work on reducing the cost of grounding
probabilistic logics: noteably, Shavlik et al
\cite{shavlik2009speeding} describe a preprocessing algorithm called
FROG that uses various heuristics to greatly reduce grounding size and
inference cost, and Niu et al \cite{niu2011tuffy} describe a more
efficient bottom-up grounding procedure that uses an RDBMS.  Other
methods that reduce grounding cost and memory usage include ``lifted''
inference methods (e.g., \cite{singla2008lifted}) and ``lazy''
inference methods (e.g., \cite{singla2006memory}); in fact, the
LazySAT inference scheme for Markov networks is broadly similar
algorithmically to PageRank-Nibble-Prove, in that it incrementally
extends a network in the course of theorem-proving.  However, there is
no theoretical analysis of the complexity of these methods, and
experiments with FROG and LazySAT suggest that they still lead to a
groundings that grow with DB size, albeit more slowly.

ProPPR is also closely related to the Path Ranking Algorithm (PRA),
learning algorithm for link prediction \cite{DBLP:journals/ml/LaoC10}.
Like ProPPR, PRA uses random-walk methods to approximate logical
inference.  However, the set of ``inference rules'' learned by PRA
corresponds roughly to a logic program in a particular form---namely,
the form
\[
\begin{array}{l}
p(S,T) \leftarrow r_{1,1}S,X_1), \ldots, r_{1,k_1}(X_{k_1-1},T). \\
p(S,T) \leftarrow r_{2,1}(S,X_1), \ldots, r_{2,k_2}(X_{k_2-1},T). \\
\vdots \\
\end{array}
\]
ProPPR allows much more general logic programs. However, unlike PRA,
we do not consider the task of searching for new logic program
clauses.

\section{CONCLUSIONS}

We described a new probabilistic first-order language which is
designed with the goal of highly efficient inference and rapid
learning.  ProPPR takes Prolog's SLD theorem-proving, extends it with
a probabilistic proof procedure, and then limits this procedure
further, by including a ``restart'' step which biases the system to
short proofs.  This means that ProPPR has a simple polynomial-time
proof procedure, based on the well-studied personalized PageRank (PPR)
method.  

Following prior work on PPR-like methods, we designed a local
grounding procedure for ProPPR, based on local partitioning methods
\cite{DBLP:journals/im/AndersenCL08}, which leads to an inference
scheme that is an order of magnitude faster that the conventional
power-iteration approach to computing PPR, takes time
$O(\frac{1}{\epsilon\alpha'})$, independent of database size.  This
ability to ``locally ground'' a query also makes it possible to
partition the weight learning task into many separate gradient
computations, one for each training example, leading to a
weight-learning method that can be easily parallelized.  In our
current implementation, an additional order-of-magnitude speedup in
learning is made possible by parallelization.  Experimentally, we
showed that ProPPR performs well, even without weight learning, on an
entity resolution task, and that supervised weight-learning improves
accuracy.

%\subsubsection*{Acknowledgements} 

%We are grateful to various numerous friends and funders, but we can't
%tell you who they are in this anonymous submission.
 
\newpage

%\subsubsection*{References} 

\bibliography{all}  
\bibliographystyle{plain}

\end{document}